%% file: main.tex
\documentclass[runningheads]{llncs}

\usepackage{eccv}

\title{TAIHRI: Task-Aware 3D Human Keypoints Localization for Close-Range Human-Robot Interaction}
\def\spaces{~~~~~~}
\author{Ao Li\inst{1,2}$^*$\spaces{}
Yonggen Ling\inst{2,3}$^*$\spaces{}
Yiyang Lin\inst{3}\spaces{}
Yuji Wang\inst{1}\spaces{} \\
Yong Deng\inst{2}\spaces{}
Yansong Tang\inst{1}$^\dag$\\
}
\institute{\textsuperscript{1}Tsinghua Shenzhen International Graduate School \\
\textsuperscript{2}Tencent Robotics X\spaces{}\textsuperscript{3}Futian Laboratory}

\usepackage{eccvabbrv}

\usepackage{graphicx}
\usepackage{booktabs}

\usepackage[accsupp]{axessibility}  

\usepackage{hyperref}

\usepackage{orcidlink}

\input{preamble.tex}

\begin{document}

\authorrunning{Li et al.}
\titlerunning{TAIHRI}

\maketitle
\begingroup
\renewcommand{\thefootnote}{}
\footnotetext{$^*$ Equal contribution.~~~~$^\dag$ Corresponding authors.}
\endgroup
\input{sec/0_abstract}    
\input{sec/1_intro}
\input{sec/2_related}
\input{sec/3_method}
\input{sec/4_experiments}
\input{sec/5_conclusion}

\section*{Acknowledgements}
This work was supported by Shenzhen Science and Technology Program (JCYJ20\\240813111903006).
%
%
\bibliographystyle{splncs04}
\bibliography{main}
\end{document}

%% file: preamble.tex
\usepackage{graphicx}
\usepackage{amsmath}
\usepackage{amssymb}
\usepackage{booktabs}

\usepackage{ctable}
\usepackage{makecell}
\usepackage{tabularx}
\usepackage{multirow}
\usepackage{arydshln}
\usepackage{afterpage}

\usepackage{wrapfig}
\usepackage{adjustbox}

\usepackage{color,xcolor}
\usepackage{colortbl}
\usepackage{enumitem}

\usepackage{pifont}
\newcommand{\cmark}{\ding{51}}  
\newcommand{\xmark}{\ding{55}}  



%% file: sec/0_abstract.tex
\begin{abstract}

Accurate 3D human keypoints localization is a critical technology enabling robots to achieve natural and safe physical interaction with users. Conventional 3D human keypoints estimation methods primarily focus on the whole-body reconstruction quality relative to the root joint. However, in practical human-robot interaction (HRI) scenarios, robots are more concerned with the precise metric-scale spatial localization of task-relevant body parts under the egocentric camera 3D coordinate. We propose TAIHRI, the first Vision-Language Model (VLM) tailored for close-range HRI perception, capable of understanding users' motion commands and directing the robot's attention to the most task-relevant keypoints. By quantizing 3D keypoints into a finite interaction space, TAIHRI precisely localize the 3D spatial coordinates of critical body parts by 2D keypoint reasoning via next token prediction, and seamlessly adapt to downstream tasks such as natural language control or global space human mesh recovery.  Experiments on egocentric interaction benchmarks demonstrate that TAIHRI achieves superior estimation accuracy for task-critical body parts. We believe TAIHRI opens new research avenues in the field of embodied human-robot interaction. Code is available at: \url{https://github.com/Tencent/TAIHRI}.

\end{abstract}

%% file: sec/1_intro.tex
\section{Introduction}
\label{sec:intro}
\input{modules/fig_teaser.tex}
Human-Robot Interaction (HRI) is a rapidly evolving research field that aims to enable seamless, natural, and intuitive interactions between humans and robotic systems across a wide range of real-world applications, including collaborative manufacturing, service robotics, and assistive technologies. A fundamental requirement for effective HRI is the robot’s ability to accurately perceive and interpret human poses, motions, and underlying intentions, which frequently depends on precise three-dimensional (3D) localization of human body keypoints. Reliable 3D keypoint estimation allows robots to reason about human spatial configurations, anticipate actions, and safely coordinate physical interactions. However, in close-range egocentric scenarios—where cameras are mounted on the robot or worn by the user—the visual observations are often subject to severe occlusion, truncation, and rapidly changing viewpoints. These factors substantially increase the difficulty of accurate 3D keypoint localization, making robust perception particularly challenging in settings that demand high precision and real-time responsiveness.

Despite substantial progress in 3D human pose and shape estimation, as demonstrated by methods such as~\cite{hmr, hmr2, pare, slahmr, camerahmr, multihmr, prompthmr, sam3dbody}, most existing approaches primarily optimize whole-body reconstruction accuracy with respect to a predefined root joint and are typically evaluated in a root-centric coordinate system. Although this formulation is effective for general pose estimation benchmarks, it does not fully satisfy the practical demands of human–robot interaction (HRI), where robots must localize task-relevant body parts with high spatial precision. 
For instance, accurate hand localization is essential for physical interactions such as handshaking or object handover, whereas reliable estimation of the underarm or torso region is crucial when assisting individuals during wheelchair transfers. 
Moreover, root-relative representations are inherently limited for egocentric robotic perception, in which spatial reasoning and motion planning must be conducted directly in the camera coordinate frame. 
To address metric-scale localization, several recent methods~\cite{multihmr, prompthmr, camerahmr, sam3dbody} explicitly incorporate camera intrinsics to recover global human pose in camera coordinates, enabling absolute depth and position estimation. 
Nevertheless, in close-range egocentric settings, the limited field of view frequently results in truncated or partially occluded human bodies, introducing severe depth ambiguities and reducing the reliability of global pose estimation. 
Such partial observations disproportionately affect task-critical regions, further complicating accurate spatial reasoning and degrading performance in interaction-sensitive applications.


To address these challenges, we propose TAIHRI, a novel Vision-Language Model (VLM) specifically designed for close-range HRI perception.
TAIHRI quantifies the localization of task-relevant body parts in a discretized interaction space tailored for egocentric scenarios.
By leveraging 2D keypoint reasoning through next token prediction, TAIHRI can accurately infer the 3D spatial coordinates of critical keypoints.
Furthermore, TAIHRI is adaptable to various downstream tasks, including natural language control and global space human mesh recovery, making it a versatile tool for HRI applications.
To emphasize the task-aware capability of TAIHRI, we design a set of interaction-centric prompts that guide the model's attention to the most relevant keypoints based on the specific HRI context.
Moreover, to enhance the model's robustness in close-range scenarios, we curate a comprehensive dataset that captures a wide range of egocentric views with various camera intrinsics, ensuring that TAIHRI can generalize effectively across diverse close-range HRI situations.

\input{modules/fig_fig2.tex}
Extensive experiments demonstrate TAIHRI's superior performance on egocentric close-range benchmarks, outperforming existing methods in 3D keypoint localization accuracy.
Moreover, compared with general VLMs, TAIHRI shows significant effectiveness in human-guided HRI tasks, validating its practical utility in real-world scenarios.
In addition, we conduct comprehensive ablation studies to underscore the effectiveness and efficiency of our pipeline and the refinements we have designed.
We apply TAIHRI on a real robotic platform to demonstrate its potential in real-world HRI applications, as illustrated in Figure~\ref{fig:teaser}, showcasing its ability to enhance the robot's perception and interaction capabilities in close-range scenarios.

%% file: modules/fig_teaser.tex
\begin{figure*}[t]
    \centering
    \includegraphics[width=0.99\linewidth]{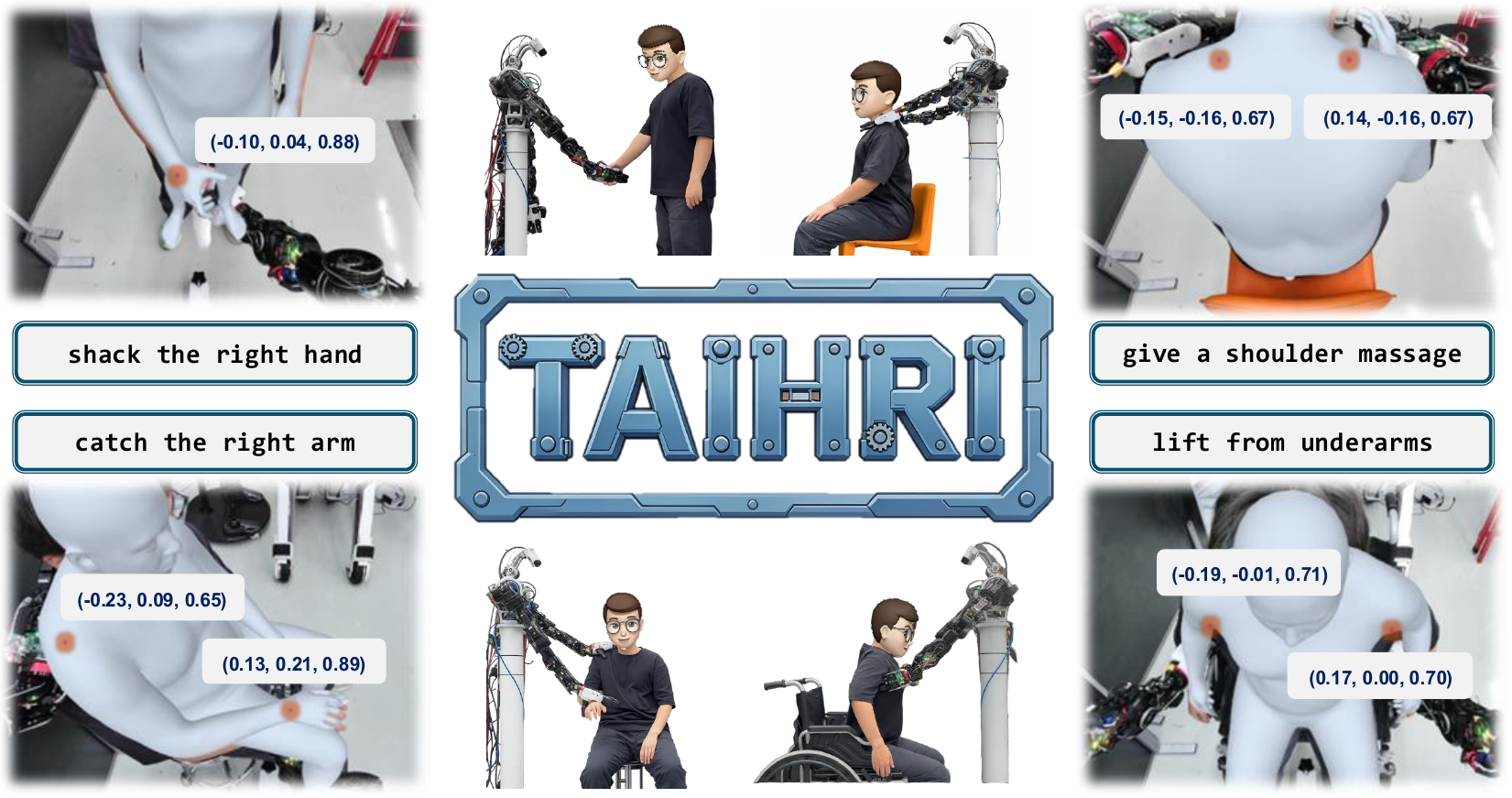}
    \vspace{-10pt}
    \caption{We introduce TAIHRI, a task-aware 3D human keypoints localization model designed for close-range human-robot interaction.}
    \label{fig:teaser}
    \vspace{-10pt}
\end{figure*}

%% file: modules/fig_fig2.tex
\begin{figure}[t]
    \centering
    \includegraphics[width=0.99\linewidth]{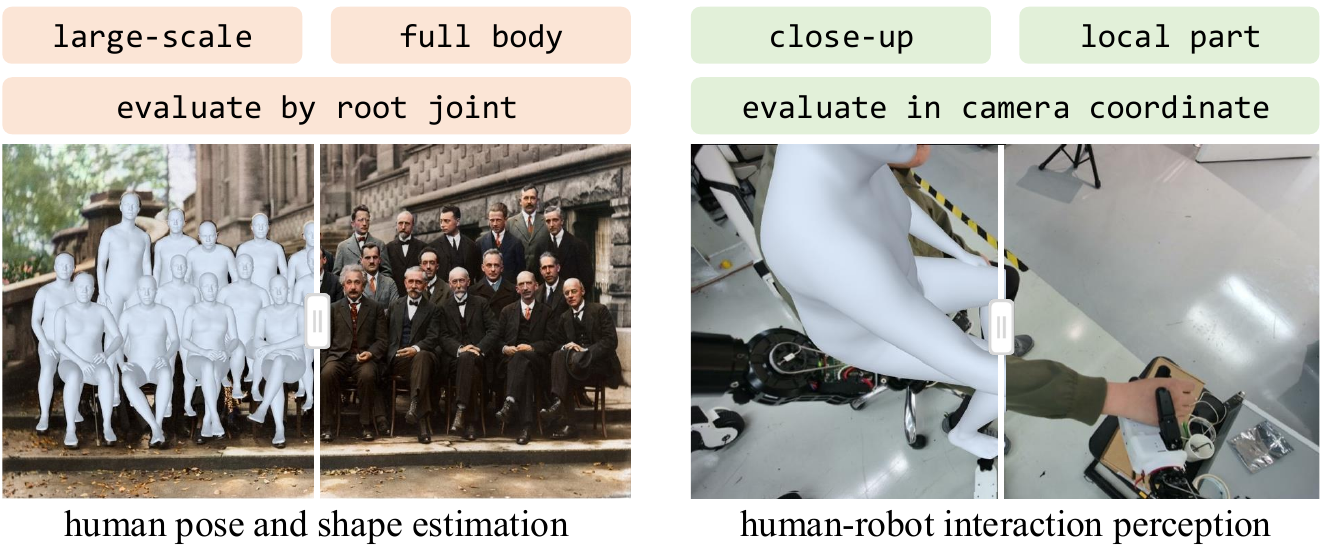}
    \vspace{-10pt}
    \caption{\textbf{The comparison of human pose and shape estimation and human-robot interaction perception.} In human-robot interaction perception, the focus is on accurately localizing task-relevant keypoints in camera space from close-range views.}
    \label{fig:fig2}
    \vspace{-15pt}
\end{figure}

%% file: sec/2_related.tex
\section{Related Work}
\label{sec:formatting}

\subsection{3D Human Pose and Shape Estimation}

Significant progress has been made in 3D human pose and shape estimation from monocular images.
SMPL model series~\cite{smpl, smplx} provide parametric representations of the human body, enabling effective modeling of human shape $\beta$ and pose $\theta$.
Early methods primarily utilized parametric models to recover full 3D human meshes from images~\cite{hmr, humor,cliff,pare}.
To enhance accuracy, subsequent approaches incorporated 2D keypoint detection as an intermediate supervision signal~\cite{choi2022learning,zhu2024dpmesh}.
Recent advancements have focused on improving model robustness in complex scenarios, such as multi-person interactions~\cite{multihmr,su2025sat}, human-object interactions~\cite{xie2022chore, xie2023visibility, li2025scorehoi} and occlusions~\cite{khirodkar2022occluded, li2023jotr}.
However, these approaches are predominantly designed for human body reconstruction in root-centric coordinates, which may not align with the egocentric perspective coordinates required for effective human-robot interaction (HRI).
To address this, methods like~\cite{prompthmr,camerahmr, sam3dbody} introduce camera intrinsics to estimate global human pose in camera coordinates.
Nevertheless, as shown in Fig.~\ref{fig:fig2}, these methods overemphasize full-body recovery and often struggle with accurate localization of task-relevant keypoints in close-range egocentric views, which are critical for human-robot interaction (HRI) applications.
Specifically, when the human subject is in close proximity to the robot-mounted camera, severe truncations and occlusions can lead to significant errors in estimating the global position of distal keypoints, such as wrists and ankles, relative to the camera.
Our TAIHRI model is specifically designed to address these challenges by directly predicting the 3D coordinates of task-relevant keypoints in camera space, thereby enhancing localization accuracy in close-range egocentric scenarios.

\subsection{VLM-based 3D Human Grounding}

Vision-Language Models (VLMs) have demonstrated remarkable capabilities in understanding and reasoning about visual content through natural language.
Early works like Flamingo~\cite{alayrac2022flamingo} and GPT-4V~\cite{gpt4v} showcased the potential of VLMs in image captioning and visual question answering.
Recently, other Multi-modal Large Language Models (MLLMs) such as GPT-5.2~\cite{gpt5}, Qwen3-VL~\cite{yang2025qwen3} and Gemini-2.5 Pro~\cite{gemini} further advanced the field by integrating large-scale vision and language pretraining, enabling more complex reasoning tasks.
Based on these foundations, several works~\cite{rexomni, wang2025vg, wang2025sam2} have explored the application of VLMs for various detailed 2D grounding tasks by next token prediction or tool learning.
To enhance the 3D understanding capabilities of VLMs, several approaches~\cite{Wang_2025_N3DVLM,Man_2025_LocateAnything3D,Li_2025_CVPR_SeeGround} have explored grounding 3D information from 2D images using from 2D cues reasoning.
However, these models are primarily designed for localizing the 3D bounding boxes of objects in general scenes, and they often lack the precision required for localizing specific human body keypoints in close-range egocentric views.
Some recent works~\cite{feng2024chatpose,feng2025posellava,lin2025chathuman} have attempted to extend VLMs for 3D human understanding and pose estimation, but they still focus on root-relative pose estimation and do not adequately release the potential of VLMs for task-aware keypoint localization.
In contrast, our TAIHRI model leverages the strengths of VLMs by next token prediction to directly predict the 3D coordinates of task-relevant human keypoints in camera space, specifically tailored for close-range egocentric HRI scenarios.

%% file: sec/3_method.tex
\section{Method}
\label{sec:method}

\subsection{Task Formulation}

Given an egocentric RGB image $I$ captured by a monocular robot-mounted camera and a human prompt $P$ that specifies the interaction task or valuable keypoints, our objective is to accurately localize a set of task-relevant 3D human keypoints $\{J_i\}_{i=1}^N$ in the robot's camera coordinate system.
These $N$ keypoints correspond to specific body parts that are critical for the intended human-robot interaction task, such as the left wrist and the left elbow when the person is raising his left hand to interact with the robot.
For specific human-robot interaction scenarios, we assume that the camera intrinsics $K_{int}$ including the focal length and the principal point are known during inference.

\input{modules/fig_dataset.tex}

\subsection{CloseHRI Dataset}
\label{sec:closehri}

Traditional 3D human pose datasets, such as Human3.6M~\cite{h36m} and MPI-INF-3DHP~\cite{3dhp}, predominantly capture full-body poses from third-person viewpoints, while datasets including AGORA~\cite{agora} and BEDLAM~\cite{bedlam,bedlamv2} provide rich multi-person scenarios but still lack the close-range egocentric perspectives essential for human–robot interaction (HRI) applications. 
To bridge this gap, we construct the CloseHRI dataset, specifically tailored for training models in close-range egocentric human–robot interaction settings, following the dataset generation pipeline of WildHuman~\cite{ge20243d} as illustrated in Figure~\ref{fig:dataset}. 
We first sample motion sequences from AMASS and render normal maps using the SMPL-X model within Blender to produce diverse human poses. 
To emulate close-range egocentric viewpoints, we place a virtual camera at distances ranging from $0.5$ to $3$ meters, with randomized height and orientation, approximating the perspectives of cameras mounted on humanoid robots. 
Using the rendered normal images as control signals, we then adopt SDXL to synthesize photorealistic images with complex backgrounds under diverse text prompts (e.g., \textit{"a person wearing a red shirt standing in a living room"}). Finally, we apply SAM3 to remove unsatisfactory samples (IoU $< 0.9$) and compute 2D keypoint reprojection errors using VitPose, retaining only samples with errors below 15 pixels for training. The resulting CloseHRI dataset contains over $1$ million images depicting diverse human subjects in varied poses and interaction scenarios, all captured from close-range egocentric viewpoints. 
This large-scale and systematically filtered dataset provides high-quality supervision for learning robust pose and interaction representations under close-range perspective views, thereby facilitating improved generalization to real-world HRI scenarios.

\subsection{Discretized Interaction Space Representation}
To seamlessly integrate with Vision-Language Models (VLMs) that excel in discrete token prediction, we discretize the continuous 3D space into a structured grid of voxels.
We assume that the robot operates within a predefined interaction volume, typically a cuboid region in front of the robot, defined by its width $W$, height $H$, and depth $D$.
Following~\cite{yang2025qwen3, chen2021pix2seq, rexomni}, we quantize the coordinates to values between $0$ and $999$ along each axis.
Therefore, the localization $(x_i, y_i, z_i)$ of each keypoint $J_i$ is converted as follows:
{
    \small
    \begin{equation*}
        \{ X_i, Y_i, Z_i \} = \left\{ \left\lfloor \frac{x_i}{W} \times 1000 \right\rfloor, \left\lfloor \frac{y_i}{H} \times 1000 \right\rfloor, \left\lfloor \frac{z_i}{D} \times 1000 \right\rfloor \right\}
    \end{equation*}
}
where $(X_i, Y_i, Z_i)$ are the discretized voxel token indices.

However, dispite the advantages of discretization, directly predicting 3D coordinates with depth dimension from a monocular image remains challenging due to the inherent ambiguity of depth estimation from a single view.
Previous works~\cite{multihmr,prompthmr,sam3dbody} introduce camera intrinsics to estimate global human pose in camera coordinates, but they require to be trained with various camera parameters and diverse datasets to generalize well.
To address this challenge, we follow~\cite{depthlm}, who claims that  intrinsic-conditioned augmentation is more effective than creating an adapter for camera parameters.
Specifically, we unify the focal length to a fixed value~(\textit{e.g.,} 1000) and adjust the image resolution accordingly during training and inference.
For the principle point, we apply the random crop augmentation to simulate the offset of the principle point, so that the model can generalize well to different image shape during inference.

\input{modules/fig_pipeline.tex}

\subsection{Thinking with 2D Keypoint Reasoning}

To effectively infer the 3D coordinates of task-relevant keypoints from monocular images, we leverage the cues of 2D keypoint locations and their spatial relationships.
Inspired by the idea of chain-of-thought prompting~\cite{wei2022chain}, we decompose the 3D keypoint localization task into two sequential sub-tasks: first predicting the 2D keypoint locations in the image plane, followed by reasoning about their corresponding depth values to obtain the final 3D coordinates.
This two-step reasoning process allows the model to first focus on accurately identifying the 2D projections of the keypoints, which is a more tractable problem given the visual information in the image.

\subsection{Reinforcement Finetuning}

Starting from an SFT-initialized model, we further optimize it with reinforcement learning using a pose-aware reward computed on visible joints $\mathcal{V}$. 
We apply the Group Relative Policy Optimization (GRPO)~\cite{shao2024deepseekmath} to finetune the model.
Given a prompt $x$, the current policy $\pi_\theta$ samples a group of $K$ responses $\{y_i\}_{i=1}^K \sim \pi_\theta(\cdot|x)$. 
Each response $y_i$ is evaluated by the pose-aware reward function $r_i = r(x, y_i)$. 
Let $\bar{r} = \frac{1}{K}\sum_{i=1}^K r_i$ denote the group mean reward, and define the group-relative advantage as $\hat{A}_i = r_i - \bar{r}$. 
The GRPO objective is defined as:
{\small
    \begin{equation*}
\begin{aligned}
\mathcal{J}_{\text{GRPO}}(\theta)
= \frac{1}{G} \sum_{i=1}^{G} \frac{1}{|o_i|} \sum_{t=1}^{|o_i|}
\Big[ \min \big( \rho_{i,t} \hat{A}_{i,t},\, \text{clip}(\rho_{i,t}, 1-\epsilon, 1+\epsilon)\hat{A}_{i,t} \big) 
- \beta \, D_{\mathrm{KL}}\!\left[ \pi_\theta \,\|\, \pi_{\text{ref}} \right] \Big].
\end{aligned}
\end{equation*}}
where $\pi_{\theta_{\text{old}}}$ denotes the behavior policy used to generate the samples, and $\epsilon$ is the clipping threshold.

For 3D joints $\mathbf{x}\in\mathbb{R}^3$ and 2D keypoints $\mathbf{u}\in\mathbb{R}^2$, we define per-joint errors $d_j=\|\hat{\mathbf{y}}_j-\mathbf{y}^{gt}_j\|_2$ (with $\mathbf{y}\in\{\mathbf{x},\mathbf{u}\}$). We use a robust Huber aggregation~\cite{meyer2021alternative} and combine it with a PCK-style success term, forming the final reward:
{\small
\begin{gather*}
E=\frac{1}{|\mathcal{V}|}\sum_{j\in\mathcal{V}}\rho_\delta(d_j), \\
r=\lambda \exp\!\left(-\frac{E}{\tau}\right)+(1-\lambda)\frac{1}{|\mathcal{V}|}\sum_{j\in\mathcal{V}}\mathbb{I}(d_j<\kappa).
\end{gather*}
}
where $\rho_\delta(d)=\frac{1}{2}d^2$ if $d\le\delta$ and $\rho_\delta(d)=\delta(d-\frac{1}{2}\delta)$ otherwise. We instantiate the same formulation for both 3D and 2D rewards by choosing the corresponding $d_j$ and consistent units for $(\delta,\kappa,\tau)$.

%% file: modules/fig_dataset.tex
\begin{figure*}[t]
    \centering
    \includegraphics[width=0.99\linewidth]{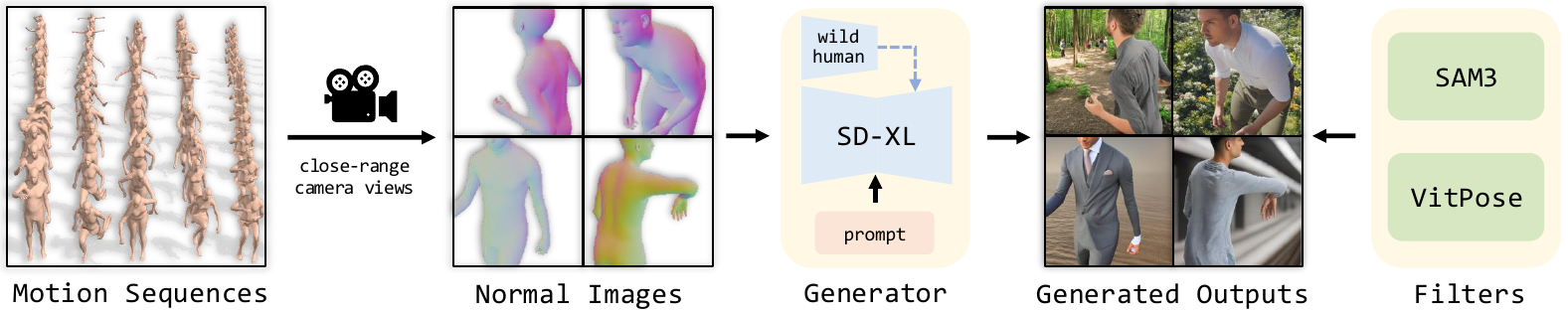}
    \vspace{-10pt}
    \caption{\textbf{The overview of CloseHRI dataset creation process.} We sample various viewpoints around the human subjects in close proximity to obtain the normal maps. Then we utilize WildHuman generator pipeline to synthesize photorealistic images and remove low-quality samples with SAM3 and 2D keypoint reprojection error filtering.}
    \label{fig:dataset}
    \vspace{-10pt}
\end{figure*}

%% file: modules/fig_pipeline.tex
\begin{figure*}[t]
    \centering
    \includegraphics[width=0.99\linewidth]{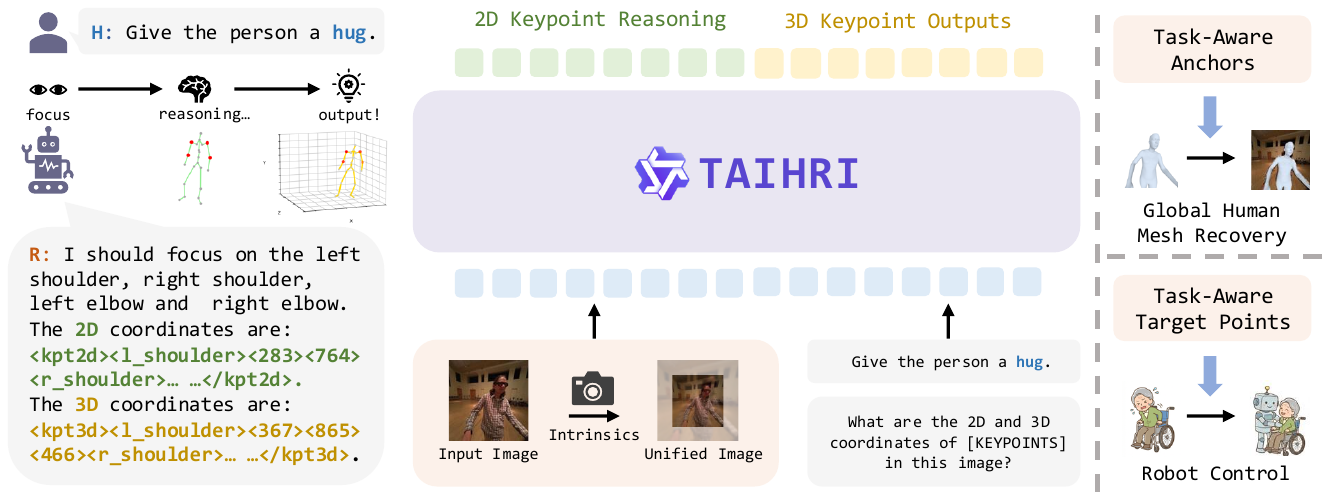}
    \vspace{-10pt}
    \caption{\textbf{The inference procedure of TAIHRI.} Given an egocentric image with the camera intrinsics and a user instruction, we first unify the focal length by image resizing. Then the image and instruction are encoded and fused by the Vision-Language Model backbone. TAIHRI will autoregressively tell the most task-relavant keypoints names, their 2D locations in pixel-level, and their 3D locations in voxel-level in the discretized interaction space. Finally, we decode the voxel-level 3D locations into metric-scale coordinates in camera space. The predicted 3D keypoints can be directly used for human-robot interaction perception or further applied to downstream tasks such as global space human mesh recovery or robot control.}
    \label{fig:pipeline}
    \vspace{-10pt}
\end{figure*}

%% file: sec/4_experiments.tex
\section{Experiment}

\subsection{Datasets and Metrics}

We train our model on the proposed CloseHRI dataset described in Section~\ref{sec:closehri}, which contains over 1 million images of diverse human subjects in various poses and interactions, captured from close-range egocentric perspectives. To enhance the model's generalization capabilities, we also incorporate additional training data from existing 3D human pose datasets, including and BEDLAMv1~\cite{bedlam}, BEDLAMv2~\cite{bedlamv2} and PDhuman~\cite{wang2023zolly}. Since we only focus on close-range egocentric scenarios, we filter out the samples where the average depth of all keypoints is larger than 3 meters. 
Finally, our training set consists of approximately 1.2 million images with diverse human poses and interactions.
We also create a prompt bank that contains over 6,000 unique interaction-centric prompts described in Section~\ref{sec:downstream} to enable natural language control during training.

We evaluate our proposed TAIHRI model on two first-view egocentric interaction datasets: Harmony4D-Test~\cite{harmony4d} and EgoBody~\cite{egobody}.
Harmony4D-Test contains 6,389 close human-human interaction frames captured from a head-mounted camera, featuring diverse human poses and interactions with contacts.
20 external cameras are used to obtain accurate 3D human keypoint annotations via multi-view triangulation.
We also evaluate on the EgoBody test set, which includes  62,155 frames of egocentric human interactions with detailed 3D body shape and pose annotations captured from 4 external cameras. 
For EgoBody test set, we sample 5,000 close-range frames for final evaluation within 3 meters.

To evaluate whether the proposed model can selectively attend to different semantically meaningful body regions, we conduct a series of controlled experiments using multiple body-part compositions designed to isolate distinct anatomical structures and interaction-relevant areas. 
Specifically, we consider four representative configurations:  \textbf{(1) Upper body}, consisting of both shoulders and both elbows; \textbf{(2) Lower body}, consisting of both hips and both knees; \textbf{(3) L-Upper}, consisting of the left shoulder, left elbow, and left wrist; and \textbf{(4) R-Upper}, consisting of the right shoulder, right elbow, and right wrist. 
These configurations are intentionally constructed to cover both symmetric (upper vs. lower body) and asymmetric (left vs. right upper limb) groupings, thereby enabling a comprehensive assessment of whether the model can adapt its attention and representation capacity to different spatial distributions and kinematic dependencies. By testing these distinct part groupings, we aim to examine the consistency, robustness, and regional sensitivity of the model in capturing informative body cues across diverse anatomical structures. 

For the evaluation metric, we employ the Global Coordinate Mean Per Joint Position Error (G-MPJPE), measured in millimeters, as the primary criterion to quantify 3D keypoint localization accuracy in metric space. Concretely, G-MPJPE computes the average Euclidean distance between predicted and ground-truth joint positions directly in the camera coordinate frame, without root alignment or rigid transformation, thereby penalizing both relative structural errors and absolute translation and depth inaccuracies. Such absolute error measurement is particularly critical for interaction-aware robotic applications, where small metric deviations may lead to unsafe contact or inaccurate motion planning. Notably, different from prior works~\cite{camerahmr, prompthmr, sam3dbody} that report MPJPE in root-relative coordinates, we evaluate errors in the global camera coordinate system to better reflect real-world HRI requirements, which demand precise metric localization within a shared physical workspace.

\subsection{Implementation Details}

We implement our TAIHRI model based on the Qwen3-VL~\cite{yang2025qwen3} 4B and 2B architectures. For the SFT stage, we use a batch size of 4 and train the model for 5 epochs with a learning rate of 2e-5. During the SFT stage, we randomly sample interaction-centric prompts from our prompt bank for each training image at 50\% probability to enhance the model's ability to understand diverse HRI contexts. The other 50\% of the time, we use a default prompt that instructs the model to localize 1 to 6 keypoints to ensure comprehensive keypoint understanding. 
For the RFT stage, we set the rollout size to 8, the KL the KL penalty coefficient $\beta$ to 0.01, and use a batch size of 32. We train the model for 5 epochs with a learning rate of 1e-6. 
We sample 10,000 images from the training set during RFT.
All experiments are conducted on 4 NVIDIA H20 GPUs.

\input{modules/tab_maintab.tex}

\subsection{Comparative Study}

\noindent
\textbf{Comparison with Global 3D Human Pose Estimation Methods}
We compare our proposed TAIHRI model with several state-of-the-art 3D human pose estimation methods that are capable of estimating the human pose with spatial translation in camera coordinates, including CameraHMR~\cite{camerahmr}, PromptHMR~\cite{prompthmr}, SAM 3D Body~\cite{sam3dbody}. 
All of the compared methods are adaptive to various camera intrinsics during inference.
As shown in Table~\ref{tab:harmony4d_egobody}, our TAIHRI model significantly outperforms all the compared methods on both Harmony4D-Egocentric and EgoBody test sets across all body-part configurations.
Visualization results in Figure~\ref{fig:main_img} further demonstrate that the joints which are far from the root point, such as wrists and ankles, are particularly challenging for previous methods to localize accurately due to the error in both pose and translation estimation. Our TAIHRI model effectively mitigates this issue by directly predicting the 3D keypoint locations in camera coordinates, leading to more precise localization of task-relevant keypoints in close-range egocentric views.

\input{modules/tab_mllm.tex}

\noindent
\textbf{Comparison with Visual-Language Models}
To further validate the effectiveness of our proposed TAIHRI model, we compare it with several state-of-the-art visual-language models (VLMs) that have demonstrated strong capabilities in 3D visual understanding and reasoning, including GPT-5.2~\cite{gpt5}, Qwen3-VL~\cite{yang2025qwen3} and Gemini-2.5 Pro~\cite{gemini}, as shown in Table~\ref{tab:mllm}. These models represent powerful general-purpose multimodal systems but are not explicitly designed for metric 3D human localization in close-range HRI settings.
We sample 50 typical test samples from the Harmony4D-Egocentric test set and evaluate their performance on the 4 Upper body keypoints localization task.
Different from the human instructions provided for TAIHRI, we guide the VLMs with the following prompt: \textit{"Given the input image, please provide the 2D and 3D coordinates of the following keypoints of the person: left shoulder, right shoulder, left elbow, right elbow. Each 2D keypoint should have x, y values in pixel level. Each 3D keypoint should have x, y, and z values in mm."}
Models like GPT-5.2 do not natively support 3D coordinate prediction from monocular image, which limits their ability to produce physically consistent outputs. Gemini-2.5 Pro outperforms other VLMs, but its predictions remain insufficiently accurate and stable for real-world robot control applications.
Our TAIHRI model achieves the best performance with a large margin and is inherently enabled for task-aware natural language control.

\input{modules/fig_main.tex}

\noindent
\textbf{Comparison with Visual Foundation Models}
We further compare our TAIHRI model with several state-of-the-art multi-modal large models and depth estimation approaches to evaluate its effectiveness in metric 3D human localization. In particular, Rex-Omni~\cite{rexomni} is designed to handle diverse 2D grounding and vision–language tasks through next-token prediction, demonstrating strong performance in semantic alignment and visual reasoning. However, as it is primarily optimized for 2D spatial understanding, it lacks explicit 3D geometric modeling and metric-scale reasoning capabilities, which limits its applicability to accurate 3D keypoint localization. For depth-based baselines, we adopt a two-stage pipeline. We first detect 2D human keypoints using VitPose~\cite{vitpose}, and then extract the corresponding depth values at those pixel locations from the predicted depth maps produced by each depth estimation model. The 2D keypoints are subsequently back-projected into 3D space using the known camera intrinsics to obtain camera-coordinate joint positions. To reduce systematic bias caused by the discrepancy between visible surface points and anatomical joint centers, we assume a fixed 2 cm offset between joints and surface depth measurements and eliminate this constant error during evaluation. As shown in Table~\ref{tab:mllm}, depth estimation methods perform poorly on the 3D keypoint localization task. This is mainly because monocular depth estimators predict per-pixel depth independently, without explicit modeling of articulated human structure or reasoning about self-occluded body parts. In contrast, our TAIHRI model jointly models human body structure and global spatial context, enabling robust inference even under severe occlusion and truncation. As a result, TAIHRI significantly outperforms all compared visual foundation models and depth-based pipelines, demonstrating superior robustness and accuracy for close-range HRI scenarios.

\input{modules/tab_abla.tex}
\subsection{Ablation Study}

\noindent
\textbf{Effectiveness of Camera Intrinsics Injection} To validate the effectiveness of injecting camera intrinsics information into our TAIHRI model, we conduct ablation studies by removing the camera intrinsics input or replacing it with a learnable ray embedding. As shown in Table~\ref{tab:abla}, we found that the image and the camera parameters captured from Harmony4D is challenging for original VLMs to infer the accurate 3D spatial locations without explicit camera intrinsics input, leading to a large performance drop. However, simply replacing the camera intrinsics with a learnable ray embedding will also degrade the performance since breaking the pre-trained parameters of the VLMs. These results demonstrate the unification of camera intrinsic information is essential for accurate 3D keypoint localization in close-range egocentric scenarios. This finding highlights that leveraging geometric priors through proper camera parameter encoding is a key component for spatial reasoning in egocentric interactions.

\noindent
\textbf{Effectiveness of Training Strategy} We further investigate the impact of our proposed two-stage training strategy. As shown in Table~\ref{tab:abla}, the 2D reasoning procedure is crucial for the next token prediction model to understand the spatial relationships between 2D and 3D keypoints, and removing this stage leads to a significant performance drop. Moreover, the RFT stage further refines the model's predictions by optimizing for long-term rewards, resulting in improved localization accuracy and more stable convergence behavior. Interestingly, we find that directly applying the MSE loss for 3D keypoint regression during the RFT stage significantly degrades performance, as MSE is highly sensitive to outliers and may cause biased gradients and suboptimal policy updates. These results validate the effectiveness of our proposed training strategy in enhancing the model's capability for accurate 3D keypoint localization. The combination of supervised learning with reinforcement learning enables the model to balance both accuracy and robustness in complex close-range interaction scenarios.

\subsection{Applications}
\label{sec:downstream}
\noindent
\textbf{Natural Language Control} To enable intuitive task-driven interaction, we construct a prompt bank containing over 6,000 interaction-centric instructions (e.g., \textit{"lift the person from the wheelchair."}) that explicitly emphasize task-relevant body regions in HRI scenarios. These prompts are carefully designed to cover diverse assistive and contact-rich interactions, encouraging the model to associate linguistic semantics with specific anatomical regions. During training, prompts from this bank are sampled with a probability of 50\% and combined with generic prompts to preserve comprehensive whole-body understanding while encouraging language-conditioned regional awareness. This mixed training strategy prevents the model from overfitting to specific templates and promotes robust semantic–spatial alignment. As shown in Figure~\ref{fig:app} (Top), the model adaptively localizes different keypoint sets according to the given instruction. For example, when prompted with \textit{"assist the person to stand from left"}, the model accurately attends to the corresponding left-side parts, demonstrating effective integration of semantic guidance and 3D spatial reasoning.

\input{modules/fig_app.tex}

\noindent
\textbf{Human Mesh Recovery in Global Coordinate} Traditional human mesh recovery methods often predict human poses in a normalized or root-relative coordinate system, which may not be directly applicable for HRI that require understanding the human's task-specific position relative to the robot.
To adapt our model for global human mesh recovery, we take the output 3D keypoints as the anchor points to align the recovered human mesh in the robot's camera coordinate system.
Given the predicted human mesh $M$ in a normalized space and the set of predicted 3D keypoints $\{K_i\}_{i=1}^N$, we compute a transformation matrix $T$ that aligns the mesh's keypoints with the predicted keypoints.
This transformation is then applied to the entire mesh to obtain its global position and orientation relative to the robot, so that the global human mesh $M_{\text{global}}$ can be expressed as $M_{\text{global}} = T(M)$.
As shown in Figure~\ref{fig:app} (Bottom), by utilizing the accurately localized 3D keypoints as anchor points, we can reposition the human mesh reconstructed from popular human mesh recovery methods~\cite{sam3dbody} in the global coordinate system. Taking "Shacking hands" as an example, we provide 1 to 3 anchor points on right arm to align the human mesh with the robot's camera system, and we evaluate the alignment quality based on the right wrist. The results demonstrate that all of the three anchor points combinations can effectively improve the accuracy compared to the root-relative baseline. 

\noindent
\textbf{Task-Aware Targets for Robot Control} As illustrated in Figure~\ref{fig:teaser}, we further develop a simple closed-loop control system for a bimanual robot to execute representative human–robot interaction tasks, such as “shaking hands” and “shoulder massage.” An Orbbec Femto Bolt camera is mounted on the robot to capture egocentric RGB images at 720p resolution, and is calibrated to align with the robot’s coordinate system to ensure consistent spatial transformation. The captured egocentric image, together with the interaction-centric prompt and camera intrinsics, is fed into our TAIHRI model to estimate task-relevant 3D keypoints in the camera coordinate frame, along with the root-relative human body mesh from~\cite{sam3dbody}, which is further aligned to the global space using our predicted anchors. For each task, we predefine corresponding human affordance regions and retarget the robot’s end-effectors to the predicted body parts through inverse kinematics, forming a perception–action loop. Experimental results demonstrate that the proposed TAIHRI model provides sufficiently accurate and stable 3D keypoint localization to guide the robot in performing precise and reliable physical interactions.

%% file: modules/tab_maintab.tex
\begin{table*}[t]
\centering
\setlength{\tabcolsep}{6pt}
\renewcommand{\arraystretch}{1.15}
\caption{\textbf{Quantitative comparison on Harmony4D-Egocentric and EgoBody test samples.} We report the G-MPJPE in millimeters for different body-part configurations. For SAM 3D Body~(3DB), we report the results of ViT backbone 3DB-H and DINOv3 backbone 3DB-DINOv3. Our TAIHRI model outperforms all the compared methods by a large margin across all configurations on both datasets.}
\vspace{-10pt}
\label{tab:harmony4d_egobody}
\small
\resizebox{\textwidth}{!}{
\begin{tabular}{l c cccc cccc}
\toprule
& & \multicolumn{4}{c}{\textbf{Harmony4D-Egocentric}} 
& \multicolumn{4}{c}{\textbf{EgoBody Test Sample}} \\
\cmidrule(lr){3-6} \cmidrule(lr){7-10}
\textbf{Methods} 
& \textbf{Focal} 
& Upper & Lower & L-Upper & R-Upper
& Upper & Lower & L-Upper & R-Upper \\
\midrule
CameraHMR~\cite{camerahmr}      & Yes & 167.50 & 165.72 & 179.22 & 169.26 & 94.92 & 118.96 & 101.27 & 99.45 \\
PromptHMR~\cite{prompthmr}      & Yes & 158.70 & 158.44 & 158.25 & 157.18 &  84.45 & 127.26 &  91.82 &  87.48 \\
3DB-H~\cite{sam3dbody}          & Yes & 127.72 & 129.73 & 143.41 & 118.77 &  88.23 & 110.58 &  93.12 &  94.06 \\
3DB-DINOv3~\cite{sam3dbody}     & Yes & 124.91 & 127.80 & 143.13 & 123.58 & 89.87 & 107.75 & 92.36 & 93.30 \\
\midrule
TAIHRI-2B  & Yes & 97.15 & 118.82 & 119.86 & 103.16 & 85.62 & 113.94 & 93.56 & 91.37 \\
\rowcolor[gray]{0.9}
TAIHRI  & Yes & \textbf{93.83} & \textbf{114.98} & \textbf{107.81} & \textbf{98.23}
               & \textbf{75.77} & \textbf{101.58} & \textbf{81.42} & \textbf{81.27} \\
\bottomrule
\end{tabular}
}
\end{table*}
\vspace{-10pt}

%% file: modules/tab_mllm.tex
\begin{table}[t]
    \def\arraystretch{1.42}
    \renewcommand{\tabcolsep}{0.8mm}
    \caption{
    \textbf{Comparison with mainstream VLMs and vision foundation models.} VP refers to VitPose~\cite{vitpose}, providing 2D detection results for depth estimation models. The 2D and 3D columns indicate whether the model can predict 2D and 3D keypoints.
    }
    \vspace{-10pt}
    \small
    \begin{center}
    \resizebox{0.75\columnwidth}{!}{
        \begin{tabular}
            {>{\raggedright\arraybackslash}m{5.30cm}|>{\centering\arraybackslash}m{0.9cm}>{\centering\arraybackslash}m{0.9cm}>{\centering\arraybackslash}m{1.75cm}}
            \specialrule{.1em}{.05em}{0.0em}
            Methods & 2D & 3D & G-MPJPE \\ 
            \hline
            \textit{* Vision Language Models} \\
            GPT-5.2~\cite{gpt5} & \cmark & \xmark & - \\            
            Qwen3-VL-235B-A22B-Instruct~\cite{yang2025qwen3} & \cmark & \cmark & 1298.3 \\
            Gemini-2.5-Pro~\cite{gemini}  & \cmark & \cmark & 436.9 \\
            \hline
            \textit{* Vision Foundation Models} \\
            Rex-Omni~\cite{rexomni}  & \cmark & \xmark & - \\
            VP + Depth Anything 3~\cite{depthanything3} & \cmark & \cmark & 352.2   \\
            VP + DepthLM~\cite{depthlm}  & \cmark & \cmark & 282.3  \\
            \hline
            \rowcolor[gray]{0.9}
            Full model  & \cmark & \cmark & \textbf{97.2} \\ 
            \specialrule{.1em}{-0.05em}{-0.05em}
        \end{tabular}
    }
    \end{center}
    \vspace{-10pt}
    \label{tab:mllm}
\end{table}

%% file: modules/fig_main.tex
\begin{figure*}[t]
    \centering
    \includegraphics[width=0.99\linewidth]{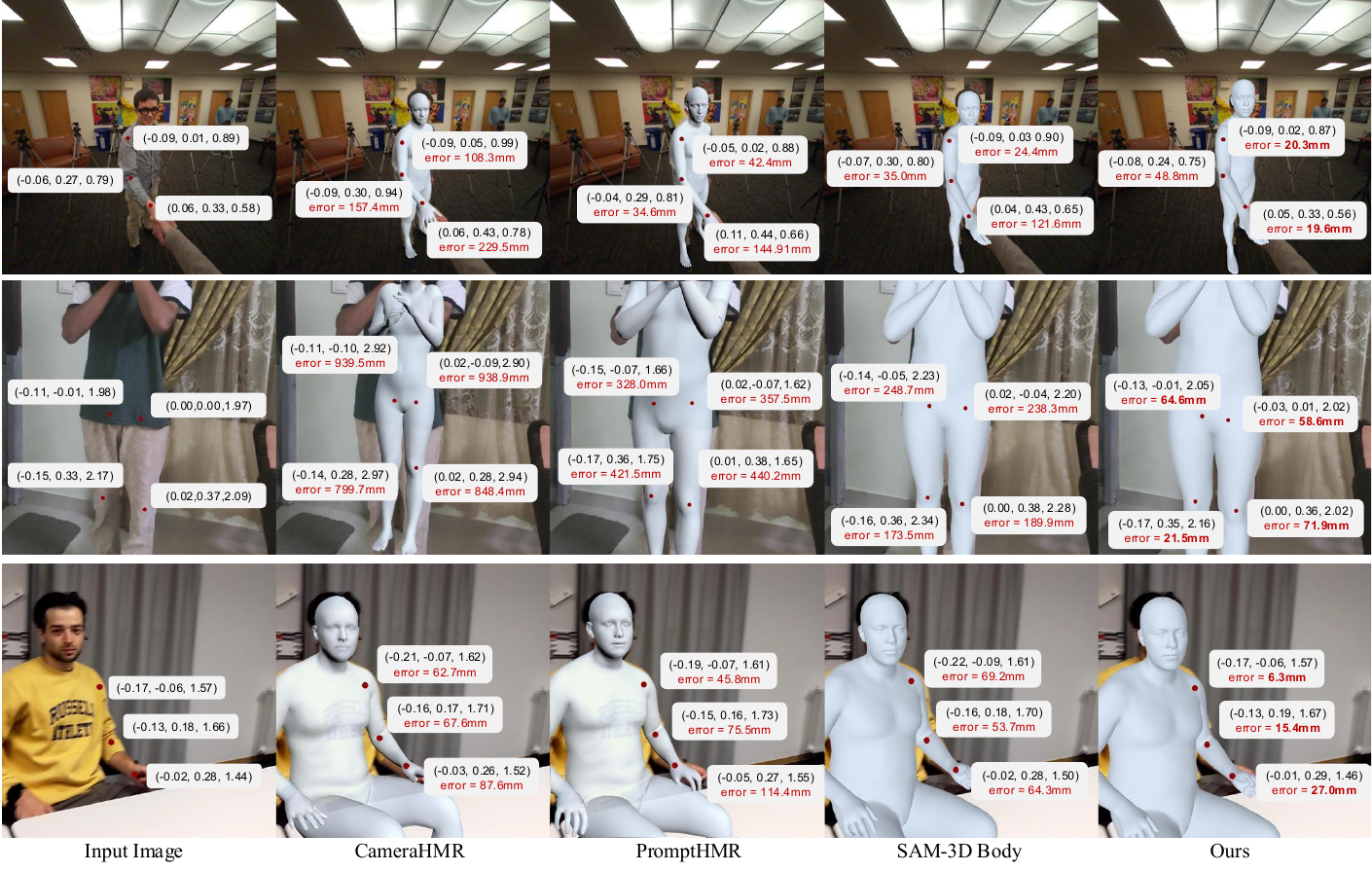}
    \vspace{-10pt}
    \caption{\textbf{The visualization comparison with SOTA methods.} We compare our proposed TAIHRI model with several state-of-the-art 3D human pose estimation methods that are capable of estimating the human pose with spatial translation in camera coordinates. Our TAIHRI precisely localizes the task-relevant keypoints, especially for those far from the root joint.}
    \label{fig:main_img}
    \vspace{-10pt}
\end{figure*}

%% file: modules/tab_abla.tex
\begin{table}[t]
    \def\arraystretch{1.42}
    \renewcommand{\tabcolsep}{0.8mm}
    \caption{
    \textbf{Ablation studies on the injection of camera intrinsics and the training strategy.} Results are reported on the Harmony4D-Egocentric test set in terms of G-MPJPE (mm). w and w/o refers to with and without, respectively.
    }
    \vspace{-10pt}
    \small
    \begin{center}
    \resizebox{0.65\columnwidth}{!}{
        \begin{tabular}
            {>{\raggedright\arraybackslash}m{2.0cm}|>{\centering\arraybackslash}m{1.2cm}>{\centering\arraybackslash}m{1.2cm}>{\centering\arraybackslash}m{1.2cm}>{\centering\arraybackslash}m{1.2cm}}
            \specialrule{.1em}{.05em}{0.0em}
            Methods & \scalebox{0.9}{Upper} & \scalebox{0.9}{Lower}  & 
            \scalebox{0.9}{L-Upper} & \scalebox{0.9}{R-Upper} \\ 
            \hline
            \textit{* Camera Int.} \\
            w/o Cam Int. & 425.13 & 433.59 & 429.47 & 428.11 \\
            w Ray Emb. & 380.29 & 423.55 & 400.09 & 382.53 \\
            \hline
            \textit{* Training} \\
            w/o 2D  & 126.67 & 138.39 & 134.50 & 126.71  \\
            w/o RFT & 101.82 & 121.26 & 110.01 & 110.79   \\
            w $r_{mse}$ & 795.24 & 831.09 & 820.75 & 798.63  \\
            \hline
            \rowcolor[gray]{0.9}
            Full model  & \textbf{93.83} & \textbf{114.98} & \textbf{107.81} & \textbf{98.23} \\ 
            \specialrule{.1em}{-0.05em}{-0.05em}
        \end{tabular}
    }
    \end{center}
    \vspace{-15pt}
    \label{tab:abla}
\end{table}

%% file: modules/fig_app.tex
\begin{figure*}[t]
    \centering
    \includegraphics[width=0.99\linewidth]{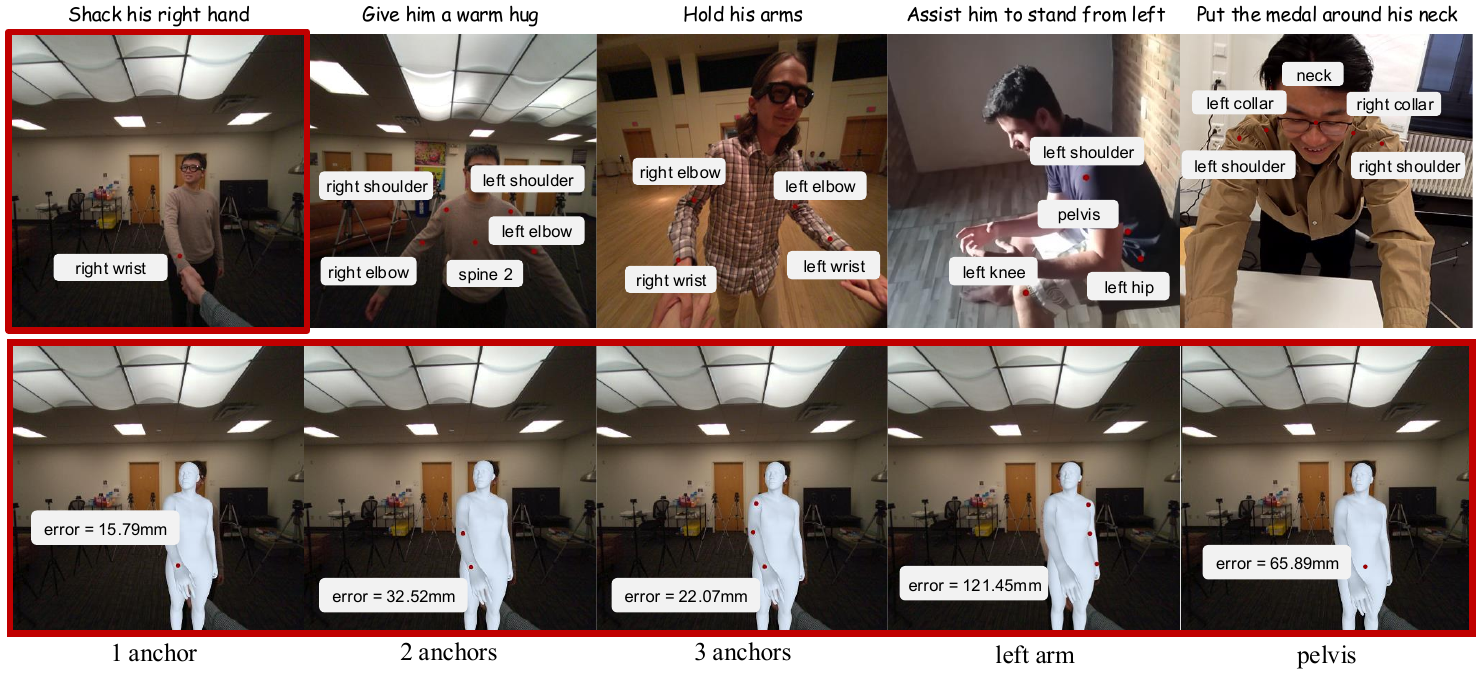}
    \vspace{-10pt}
    \caption{\textbf{The demonstration of applications.} Top: By leveraging the natural language control capability of our method, users can flexibly specify different keypoints for localization based on various interaction tasks. Bottom: Utilizing the accurately localized 3D keypoints as anchor points, we can reposition the human mesh in the global coordinate system.}
    \label{fig:app}
    \vspace{-10pt}
\end{figure*}

%% file: sec/5_conclusion.tex
\section{Conclusion}

This paper presents TAIHRI, the first Vision-Language Model specifically designed for close-range human-robot interaction perception, which bridges the gap between conventional root-centric whole-body 3D pose estimation and the practical need for egocentric, task-relevant keypoint localization. By discretizing the interaction space for 3D keypoints, leveraging 2D keypoint reasoning via next token prediction, and introducing effective task-aware prompts that guide attention to application-specific body parts in a controllable manner, TAIHRI achieves superior accuracy on egocentric interaction benchmarks and substantially outperforms prior methods on task-critical body parts, while remaining easily adaptable to various downstream tasks such as natural language control and global space human mesh recovery.
We believe these findings establish VLM-based, task-aware perception as a promising paradigm for embodied HRI and open up new directions toward more intuitive, robust, and safe human–robot collaboration in real-world environments.